\documentclass[]{article}

\usepackage{indentfirst}
\usepackage{amsthm,amsmath,amssymb}
\usepackage{mathrsfs}
\usepackage{graphicx}
\usepackage{hyperref}
\usepackage{algorithmicx}
\usepackage{algorithm}
\usepackage{algpseudocode}
\usepackage{booktabs}
\usepackage{tabularx}
\usepackage{multirow}
\usepackage{lscape}{\footnotesize }
\usepackage{ulem}
\usepackage{geometry}
\hypersetup{
	colorlinks=true,
	linkcolor=blue,
	filecolor=magenta,
	urlcolor=cyan,
}

\title{DDMT: Denoising Diffusion Mask Transformer Models for Multivariate Time Series Anomaly Detection}
\author{YangChaocheng}

\begin{document}

\maketitle

\begin{abstract}
	
Anomaly detection in multivariate time series has emerged as a crucial challenge in time series research, with significant research implications in various fields such as fraud detection, fault diagnosis, and system state estimation. Reconstruction-based models have shown promising potential in recent years for detecting anomalies in time series data. However, due to the rapid increase in data scale and dimensionality, the issues of noise and Weak Identity Mapping (WIM) during time series reconstruction have become increasingly pronounced. To address this, we introduce a novel Adaptive Dynamic Neighbor Mask (ADNM) mechanism and integrate it with the Transformer and Denoising Diffusion Model, creating a new framework for multivariate time series anomaly detection, named Denoising Diffusion Mask Transformer (DDMT). The ADNM module is introduced to mitigate information leakage between input and output features during data reconstruction, thereby alleviating the problem of WIM during reconstruction. The Denoising Diffusion Transformer (DDT) employs the Transformer as an internal neural network structure for Denoising Diffusion Model. It learns the stepwise generation process of time series data to model the probability distribution of the data, capturing normal data patterns and progressively restoring time series data by removing noise, resulting in a clear recovery of anomalies. To the best of our knowledge, this is the first model that combines Denoising Diffusion Model and the Transformer for multivariate time series anomaly detection. Experimental evaluations were conducted on five publicly available multivariate time series anomaly detection datasets. The results demonstrate that the model effectively identifies anomalies in time series data, achieving state-of-the-art performance in anomaly detection.

\end{abstract}

\providecommand{\keywords}[1]
{
	\small	
	\textbf{Keywords:} #1
}


\section{Introduction}

\setlength{\parindent}{2em}

With advancements in sensing and monitoring technologies, the operational status of modern devices can be recorded in real-time, leading to the generation and accumulation of massive amounts of multivariate time series data. Time series analysis involves modeling these data for classification, prediction, and anomaly detection.  Among them, anomaly detection is a significant research direction in time series analysis. Time Series Anomaly Detection (TSAD) refers to analyzing and modeling time series data to detect outliers or abnormal patterns within them. TSAD has been extensively studied in various application domains, including credit card fraud detection, intrusion detection in network security, and fault diagnosis in industrial settings \cite{blazquez2021review}. Effective anomaly detection can provide timely warning information, allowing personnel to gain valuable time and prevent disasters from occurring. Therefore, a timely and efficient anomaly detection system is of paramount importance.

Existing methods for TSAD can be classified into traditional methods and deep learning-based methods. Traditional methods \cite{brauckhoff2009applying,farooqi2008intrusion,huang2013lof,jianliang2009application,liu2008isolation} have shown good performance in detecting shallow-level sequence anomalies but do not consider the temporal information and are difficult to generalize to unseen real scenarios\cite{xu2021anomaly}. With the increase in data scale and complexity, deep learning methods have gradually been applied to TSAD due to their ability to extract deep complex features and handle massive amounts of data \cite{habeeb2019real}. Examples of such methods include Recurrent Neural Networks (RNN) \cite{hundman2018detecting}, Variational Autoencoders (VAE) \cite{xu2018unsupervised}, and Generative Adversarial Networks (GAN) \cite{bashar2020tanogan,geiger2020tadgan,li2019mad}. Given the presence of long-term dependencies in time series data, many researchers commonly employ models based on Long Short-Term Memory (LSTM) \cite{hochreiter1997long} or Gated Recurrent Unit (GRU) \cite{chung2014empirical} to capture the long-term dependencies in time series. The gate mechanisms of LSTM and GRU partially alleviate the vanishing gradient problem associated with long-term dependencies. However, the dependence on the previous time step's computation results in these models lacking parallelization capabilities.

In previous research, the use of generative models such as VAE \cite{xu2018unsupervised} and GAN \cite{geiger2020tadgan} has been explored for TSAD. This class of models excels in learning and simulating the probability distribution of data, thereby generating new samples similar to the training data. However, because the VAE model assumes that the latent space is continuous, while real-world data may have discrete characteristics, this could result in less precise latent representations being learned. As a result, the generated samples from VAE models are often not as realistic as those from GAN models. On the other hand, the training process of GAN models involves adversarial learning, which can lead to issues such as vanishing gradients and mode collapse, resulting in training instability\cite{wu2023decompose}. In TSAD, these models can generate similar time series based on the original distribution of the time series. However, they are also prone to fitting the anomalous portions of the time series, thus compromising the anomaly detection performance of the models.

The Denoising Diffusion Model has recently emerged as a new state-of-the-art generative model, demonstrating impressive sample quality and diversity\cite{xiao2022tackling}. Compared to GAN, the Denoising Diffusion Model offers better pattern coverage and higher sample quality than VAE \cite{wyatt2022anoddpm}. It has become a leading method in various visual tasks, such as text-to-image translation, super-resolution, and image synthesis \cite{nichol2021glide,ho2022cascaded,rombach2022high}, garnering widespread attention. The step-by-step noise removal approach employed in Denoising Diffusion Models enables the generation of high-quality samples. Moreover, its training process is more stable than GAN models, as it is less prone to issues such as training instability and mode collapse. However, in previous image modeling work, the standard architecture for $\hat{x}_{\theta}$ in diffusion models was the U-Net \cite{ho2020denoising,ronneberger2015u}. The U-Net model utilized a stack of residual layers and downsampled convolutions, followed by another stack of residual layers and upsampled convolutions. It employed skip connections between layers with the same spatial dimensions, enabling the model to effectively capture local features within image data and obtain spatial positional information for pixels. As a result, it excelled in numerous image tasks \cite{dhariwal2021diffusion}. Nevertheless, because time series contain deep local features and complex temporal dependencies, the original model architecture may prove challenging to achieve the desired results in time series research.

In recent years, the Transformer model has made significant advancements in various fields, including natural language processing \cite{brown2020language}, computer vision \cite{liu2021swin}, and time series analysis \cite{zhou2021informer}. Its success stems from its ability to capture dependencies between data elements and its powerful feature extraction capabilities. Compared to the structure of RNN, the Transformer model utilizes self-attention mechanisms, enabling adaptive modeling of both short-term and long-term dependencies through pairwise interactions between query and key. It also computes information for the entire sequence in parallel, improving computational efficiency. Therefore, we have combined the Transformer with the Denoising Diffusion Model, replacing U-Net as the internal neural network architecture within the diffusion model. This modification enables our model to better handle global dependencies, complex patterns, and long-term relationships within time series data, consequently enhancing the performance of time series reconstruction. However, when using the Transformer for time series reconstruction, each node's reconstruction relies significantly on its own information and that of neighboring nodes, making the output heavily dependent on the input. This situation can result in the reconstruction of anomalous parts during sequence reconstruction, a phenomenon we refer to as Weak Identity Mapping (WIM).

To address the issue of WIM during time series reconstruction, we propose a novel ADNM approach that leverages the differences between the reconstructed time series and the original time series to set the mask. Currently, popular reconstruction networks \cite{vaswani2017attention,chen2021learning,bashar2020tanogan}, while effective at extracting features from the original time series, may encounter difficulties during time series reconstruction if the input data contain anomalies. In such cases, the model might attempt to incorporate these anomalies into the reconstruction, resulting in inaccurate results and exacerbating the problem of WIM during time series reconstruction. Similarly, complex autoencoder structures \cite{lin2020anomaly} also possess strong feature extraction capabilities. Consequently, when computing the differences between the original time series and the reconstructed time series, the errors in the normal and anomalous portions of the data may not exhibit noticeable distinctions, which makes it challenging to utilize these errors to establish a dynamic mask.

\begin{figure}
	\centering
	\includegraphics[width=0.8\linewidth]{"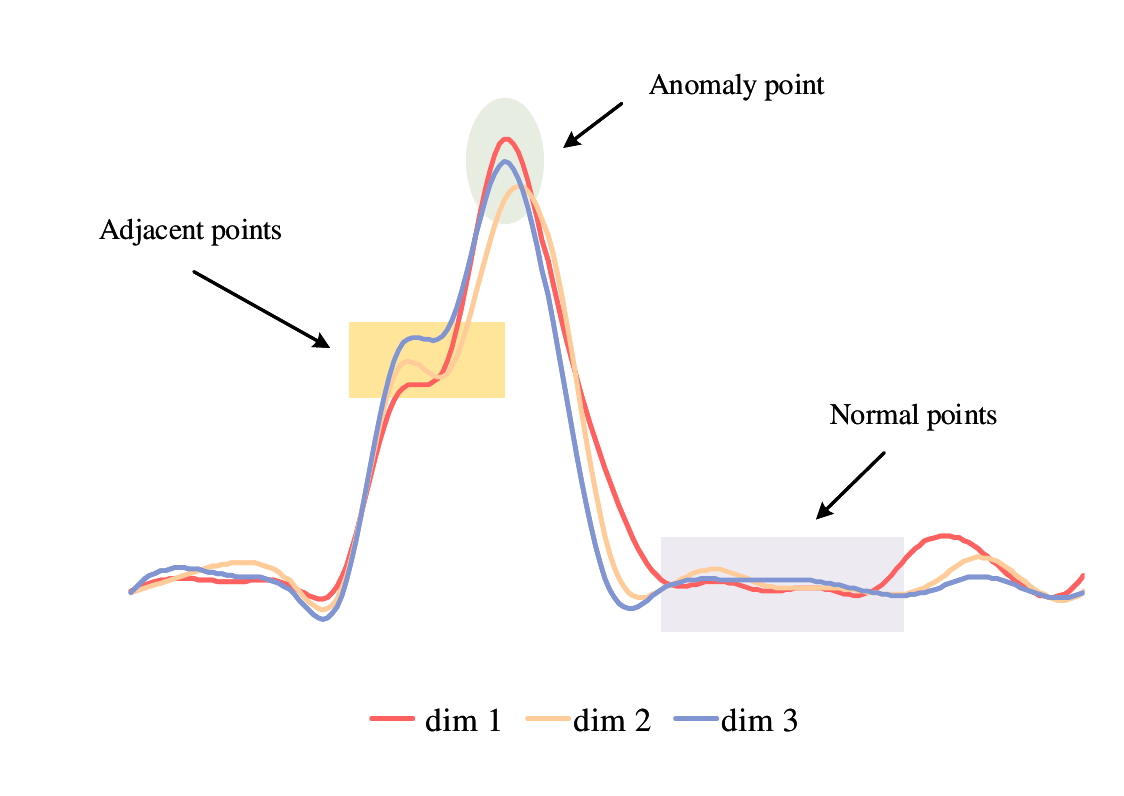"}
	\caption{The neighboring points of anomaly points may also exhibit abnormal data patterns.}
	\label{Fig.1}
\end{figure}

In light of these challenges, we have adopted a simple autoencoder structure. While it may not fully explore deep-level features within the time series, it allows us to better downplay the information related to abnormal data patterns. Consequently, during the time series reconstruction process, the differences in a small portion of normal data might appear relatively large. However, the differences in the data from the abnormal portions will be even more significant. Subsequently, we dynamically adjust the mask based on the differences between the reconstructed and original time series. It is essential to note that data points adjacent to anomalies may not exhibit the characteristics of normal data (as shown in Fig. \ref{Fig.1}). Therefore, when the model extracts local features and long-term dependencies within the time series, it might inadvertently learn some abnormal pattern features. To mitigate this, we employ the ADNM before utilizing the Transformer to compute self-attention within the time series. ADNM masks nodes with significant errors in the sequence and their neighboring nodes, preventing them from utilizing their own and neighboring information during data reconstruction. Consequently, the model must rely on information from more distant points to reconstruct the current time point. This approach enhances the model's ability to understand normal patterns and fit normal data distribution during this process.

The principal contributions of this paper are outlined as follows:
\begin{itemize}
\item 
We have designed a novel framework that combines the Transformer and Denoising Diffusion Model for TSAD. To the best of our knowledge, this is the first approach that integrates the Denoising Diffusion Model and Transformer for TSAD. It addresses global dependencies, complex patterns, and long-term relationships within time series data and progressively restores time series data by removing noise, making it easier for the model to capture and identify anomalies.
\item
We propose a new ADNM mechanism that dynamically adjusts the mask based on the differences between the reconstructed and original time series. This mechanism alleviates the problem of WIM during sequence reconstruction, enhancing the model's generalization capability.
\item 
Conducting extensive experiments on five publicly available multivariate TSAD datasets. The results demonstrate that the proposed DDMT achieves state-of-the-art performance in anomaly detection.
\end{itemize}
\section{Related Work}
In this section, a brief overview of methods for TSAD is provided, including machine learning-based methods and deep learning-based methods.
\subsection{Machine Learning-based Methods}
We categorize machine learning-based anomaly detection methods into four categories: linear-based models, distance-based models, density-based models, and classifier-based models.

One popular method in linear models is Principal Component Analysis (PCA) \cite{brauckhoff2009applying}. PCA is a multivariate data analysis technique that can extract significant variability information from the data while reducing its dimensionality. It aims to preserve the diversity of the data as much as possible. However, this method is noise-sensitive and assumes that the data follow a multivariate Gaussian distribution \cite{dai2013model}.

K-means \cite{jianliang2009application} is a distance-based clustering algorithm that aims to partition data points into different clusters. The algorithm calculates the distances between all samples and the cluster centers, and then determines whether certain points are outliers based on the compactness of the clusters. Another commonly used distance-based method is K-Nearest Neighbors (KNN) \cite{farooqi2008intrusion}. This algorithm calculates the distances between a data point and its K nearest neighbors and derives an anomaly score based on these distances. These distance-based methods often require prior knowledge about the duration of anomalies and the number of anomalies \cite{li2019mad}. Moreover, they may not capture temporal correlations \cite{geiger2020tadgan}.

Local Outlier Factor (LOF) \cite{huang2013lof,breunig2000lof} and Density Based Spatial Clustering of Applications with Noise (DBSCAN) \cite{ester1996density} are density-based anomaly detection algorithms. These algorithms compute the density difference between each data point and its surrounding neighbors and calculate an outlier factor for each data point based on this difference to identify anomalies \cite{huang2013lof}. These algorithms are suitable for large-scale datasets and can be effectively scaled and computed in parallel. However, they are only effective when the outliers in the data do not form significant clusters, making them more demanding in terms of data requirements.

Bayesian networks \cite{kruegel2003bayesian} and Support Vector Machines (SVM) \cite{xie2012intelligent,ma2003time} are two common anomaly detection algorithms based on classifier models. Anomaly detection algorithms based on Bayesian networks have been applied in areas such as network intrusion and image anomaly detection. However, this algorithm relies on prior knowledge to determine the dependencies between variables. If the prior knowledge is insufficient or inaccurate, it may result in an inaccurate network structure and affect the model's effectiveness. SVM constructs a hyperplane enclosing most of the normal samples and then uses the distance from a test sample to the boundary of the hyperplane to determine if it is an anomaly. However, its training process involves solving large-scale quadratic programming problems, which can be computationally expensive, especially for high-dimensional data, requiring significant computational time and resources.

\subsection{Deep Learning-based Methods}

Labeling anomalies in time series requires costly expert involvement and cannot guarantee coverage of all anomalies. Therefore, deep learning-based unsupervised anomaly detection methods have been widely adopted due to their excellent performance \cite{li2019mad}. Deep learning methods have the ability to learn complex dynamics within the data without making assumptions about underlying patterns, making them highly attractive choices in contemporary time series analysis \cite{choi2021deep}. For instance, autoencoders (AE) identify anomalies by calculating reconstruction errors through sequence reconstruction \cite{chen2018autoencoder}. Other methods, such as VAE \cite{xu2018unsupervised} and LSTM-based variational autoencoders \cite{park2018multimodal}, have also demonstrated strong performance. Chen et al. \cite{chen2023semisupervised} proposed a semi-supervised anomaly detection method based on variational autoencoders. They introduced a parallel multi-head attention mechanism to capture the long-term dependencies in multivariate time series, while using LSTM to capture short-term dependencies. Furthermore, anomaly detection models based on Generative Adversarial Networks (GANs) have also gained attention. Li et al. \cite{li2019mad} replaced the generator and discriminator in GANs with the structure of LSTM-RNN. They randomly generated samples in the latent space and trained the generator model to generate realistic samples. Niu et al. \cite{niu2020lstm} investigated a model that combines LSTM, VAE, and GAN for TSAD. They proposed an LSTM-based VAE-GAN method by jointly training the encoder, generator, and discriminator. However, using the LSTM structure directly within the GAN models leads to significantly high computational complexity due to the iterative process involved \cite{choi2020gan}, and it is prone to problems such as training instability and mode collapse.

Since Vaswani introduced the Transformer for translation \cite{vaswani2017attention}, many Transformer-based models have been gradually applied in time series research. Chen et al. \cite{chen2020multi} combined multi-task learning with the Transformer architecture and proposed the MTL-Trans model for time series modeling and multi-dimensional time series prediction. Yan et al. \cite{yan2022land} introduced a time series classification approach using the Informer network, which leverages both local and global dependency features of land cover time series. Wang et al. \cite{wang2022variational} proposed an unsupervised TSAD method based on the Transformer. They employed a multi-scale fusion algorithm for time series data, combining features from multiple time scales to obtain expressive representations of the time series. These studies demonstrate that Transformer-based models can better capture the long-range dependencies in time series. Therefore, we incorporate this model into anomaly detection for time series data. In contrast to the vanilla Transformer, our approach incorporates a simple yet effective masking mechanism called ADNM before performing self-attention computations with the Transformer. ADNM masks out anomalous nodes within the sequence along with their neighboring information.   This approach prevents information leakage from input features to output features, thus alleviating the problem of WIM during sequence reconstruction.

Recently, the Denoising Diffusion Model has demonstrated tremendous potential in image generation tasks and has shown remarkable performance in various domains such as computer vision, natural language processing, and audio processing. For example, Brempong et al. \cite{brempong2022decoder} proposed a denoising-based decoder pretraining method and connected a denoising autoencoder with a diffusion probabilistic model for semantic image segmentation. Li et al. \cite{li2022diffusion} developed a novel continuous diffusion-based non-autoregressive language model that enables fine-grained control in text generation. Kong et al. \cite{kong2020diffwave} introduced a universal diffusion probabilistic model called DiffWave for raw audio synthesis that is capable of generating high-quality audio signals for both conditional and unconditional waveform generation tasks. Currently, several methods based on the Denoising Diffusion Model are being used in time series research. For instance, Alcaraz et al. \cite{alcaraz2022diffusion} combined the diffusion model with structured state space models and proposed a novel method for time series imputation. Rasul et al. \cite{rasul2021autoregressive} introduced an autoregressive model called TimeGrad for multivariate probabilistic time series forecasting. It samples from the data distribution at each time step by estimating gradients and then performs predictions for multivariate time series data.

Despite the gradual application of the Denoising Diffusion Model in time series research, multivariate time series can involve multiple dimensions such as time, space, and attributes. Moreover, time series data often exhibit long-range dependencies. To address these challenges, we propose combining the Denoising Diffusion Model with the Transformer architecture to construct a generative model suitable for time series data. This integration aims to facilitate efficient TSAD, considering the complex nature of multivariate time series and their long-range dependencies.

\section{Method}
In this section, we provide a brief overview of multivariate TSAD. We then introduce the components of the DDMT and describe the workflow for anomaly detection.

\subsection{Problem Description}
We consider a multivariate time series $\mathcal{T}$=\{$x_{1}$,\ldots,$x_{T}$\}, where each element $x_i$ has $t$ timestamps, $t \in [1,T]$, and C dimensions, $c\in[1,C]$. Here, a univariate time series is a special case where $c=1$. The multivariate time series $\mathcal{T}$ can be represented as a matrix:

\begin{center}
$\mathcal{T}=\begin{pmatrix}\mathbf{x}_{(1,1|i)}&\mathbf{x}_{(2,1|i)}&...&\mathbf{x}_{(T,1|i)}\\ \mathbf{x}_{(1,2|i)}&\mathbf{x}_{(2,2|i)}&...&\mathbf{x}_{(T,2|i)}\\ \vdots&\vdots&\ddots&\vdots\\ \mathbf{x}_{(1,C|i)}&\mathbf{x}_{(2,C|i)}&...&\mathbf{x}_{(T,C|i)}\end{pmatrix}$
\end{center}

The task of anomaly detection involves providing a multivariate time series $\mathcal{T}$ and training it to discover pattern features within the time series. Subsequently, in a test set where the patterns are the same as $\mathcal{T}$, a label $\hat{y}_t$ (0 for normal and 1 for anomaly) is assigned to each time point $x_t$. This process enables the detection of abnormal instances that do not conform to the established patterns.

\subsection{Overview}

\begin{figure}
	\centering
	\includegraphics[width=1.0\linewidth]{"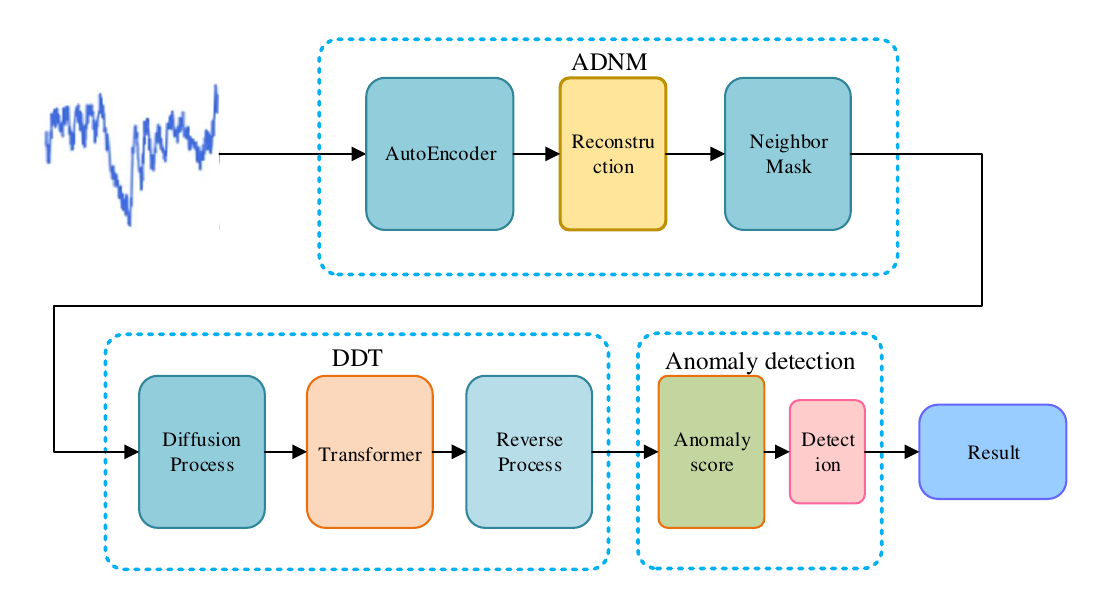"}
	\caption{The overall framework of DDMT. The ADNM module calculates errors and adjusts masks, the DDT module learns the signal mapping, and the anomaly detection module identifies abnormal nodes by comparing the generated sequence with the original sequence.}
	\label{Fig.2}
\end{figure}

The overall architecture of DDMT, as shown in Fig. \ref{Fig.2}, consists of three components: ADNM, DDT, and Anomaly Detection. The ADNM calculates the error between the reconstructed time series and the original time series and dynamically adjusts masks based on this error. DDT comprises the denoising diffusion model and the Transformer, where the Denoising Diffusion Model learns the mapping from the latent space to the signal space by removing noise that is sequentially added in a Markovian manner during the forward process, which corresponds to the noise added in the forward process \cite{kong2020diffwave}, with the Transformer serving as the internal neural network within the diffusion model. The anomaly detection component assigns an anomaly score to each node in the time series based on the differences between the generated and original sequences, ultimately assessing the anomaly status of each node in the time series.

\subsection{Adaptive Dynamic Neighbor Mask}

The traditional self-attention mechanism \cite{vaswani2017attention} computes the correlation between each position in a sequence by attending to itself and other positions, thereby capturing global contextual information. However, this characteristic also leads to the WIM issue in anomaly detection models \cite{you2022unified}. In a scenario with full attention, each node is allowed to attend to itself, enabling it to reconstruct itself through simple self-replication during sequence reconstruction. Consequently, this approach struggles to uncover deep, intricate features within the time series data.

\begin{figure}
	\centering
	\includegraphics[width=0.59\linewidth]{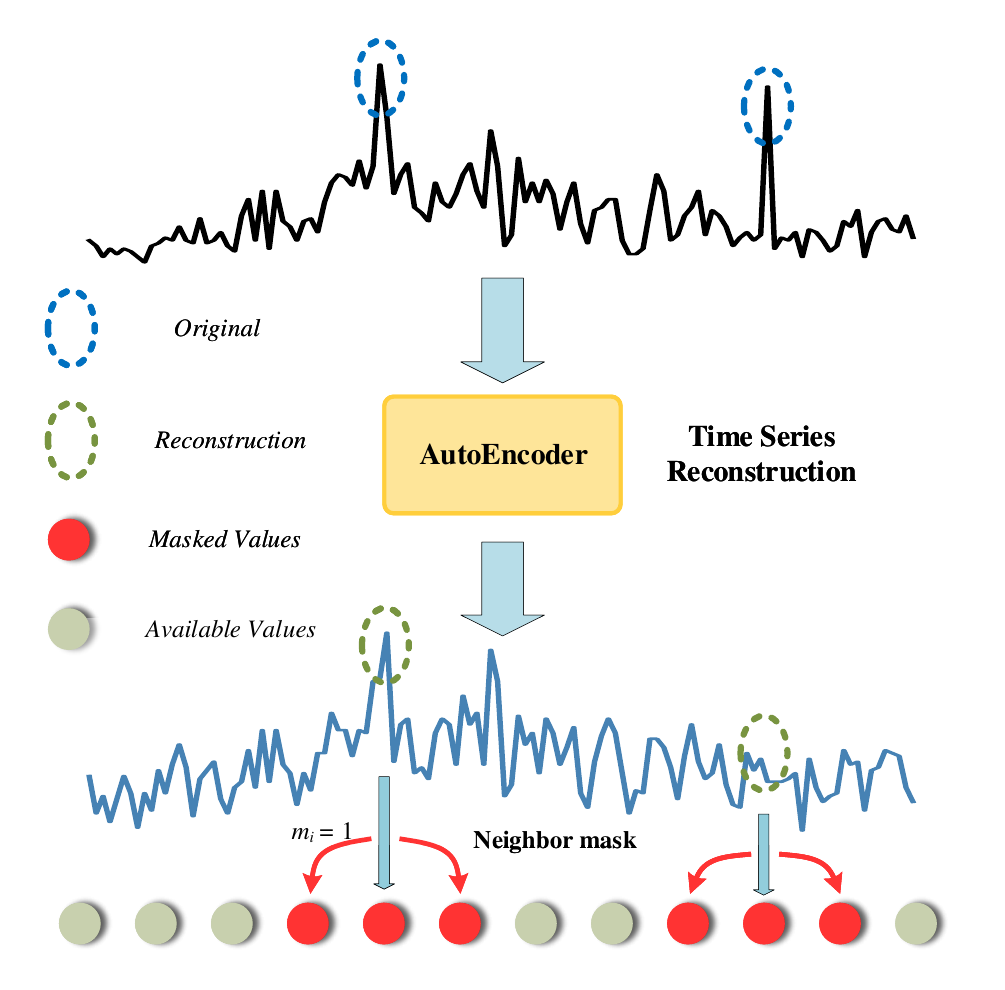}
	\caption{Adaptive Dynamic Neighbor Mask(ADNM). $m_i=1$ indicates that the number of masked neighboring nodes is 1, and this value is adjusted based on the Pearson correlation coefficient threshold. If $m_i=0$, no masking operation is performed on neighboring nodes.}
	\label{Fig.3}
\end{figure}

To address this challenge, we have devised a novel ADNM. In contrast to directly employing a self-attention mechanism, ADNM employs an Autoencoder to perform an initial reconstruction of the time series before applying the Transformer's self-attention. Subsequently, it masks nodes with significant reconstruction errors and their neighboring nodes, effectively blocking self-information and information from closely adjacent neighbors. As a result, the model can focuses more on non-adjacent regions, making it more difficult to reconstruct anomalies during the process. The Fig. \ref{Fig.3} illustrates the working process of ADNM.

For each node $i$, we introduce a mask scale $m_i\in N_i$, where $N_i$ represents the total number of neighboring nodes for node $i$, and $m_i$ is the number of neighboring nodes that need to be masked. Each node is associated with an reconstruction error, and based on this reconstruction error and the actual anomaly ratio, we can determine which points need to be masked. We then utilize the Pearson correlation coefficient between the masked node and its neighboring nodes to determine the size of the mask scale. If the correlation coefficient is greater than a certain threshold, the mask window can expand in both directions; otherwise, it stops. This approach allows mi¬ to adaptively adjust within the range $[0, N_i]$. Consequently, we obtain the mask matrix $M$. The mask matrix is dynamically generated based on the reconstruction errors of the time series and the required number of masked neighboring nodes $m_i$, which determines which nodes should be masked in the subsequent self-attention calculation process, as shown in Fig. \ref{Fig.3}.

ADNM adopts a simple autoencoder structure with the primary goal of disregarding information related to abnormal data patterns.  While this autoencoder structure may not fully explore deep-level features within the time series, it excels at minimizing the impact of anomalous data.  By selectively masking neighboring nodes, the ADNM elevates the model's attention toward informative non-neighboring nodes while reducing the influence of the self-node on sequence reconstruction.  This mechanism facilitates improved feature learning and enhances the model's generalization ability.

\subsection{Denoising Diffusion Transformer}

\begin{figure}
	\centering
	\includegraphics[height=7cm]{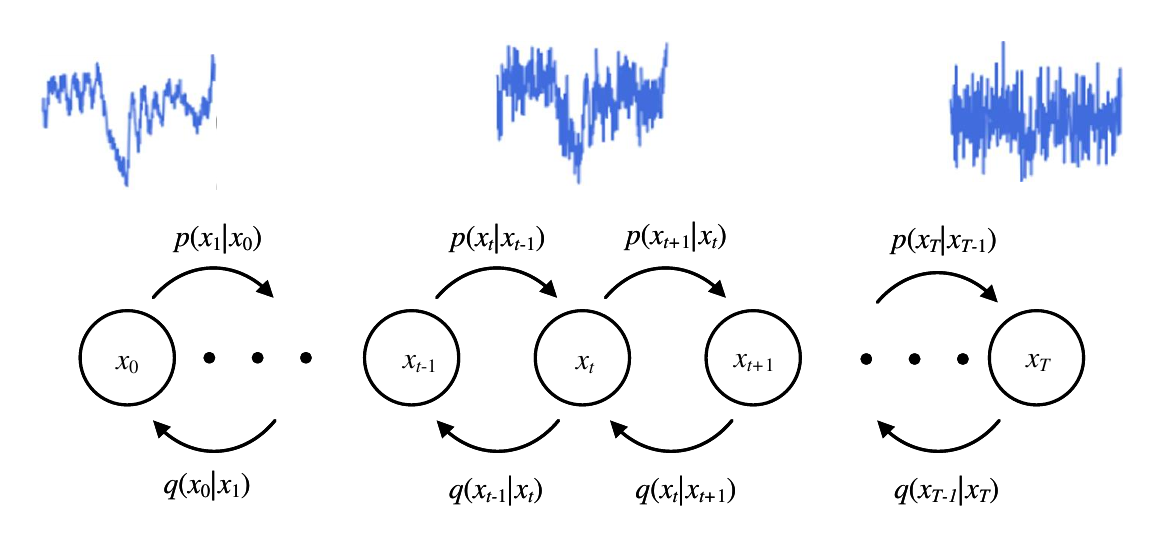}
	\caption{The forward and backward processes for Denoising Diffusion Model. The forward process makes the time series approach a Gaussian distribution gradually by adding noise, and the backward process restores the time series by removing the noise step by step.}
	\label{Fig.4}
\end{figure}

Denoising Diffusion Model represents a class of generative models that have demonstrated state-of-the-art performance on a diverse range of domains, including images \cite{dhariwal2021diffusion}, audio \cite{kong2020diffwave}, and video data \cite{ho2022video}, among others. Denoising Diffusion Model treats a stochastic process as a partial differential equation and solves this equation through numerical methods to obtain the probability distribution function of the stochastic process. As the noise gradually diffuses, the probability distribution function of the stochastic process evolves, reflecting the underlying dynamics. Denoising Diffusion Model consists of a forward diffusion process, denoted as $q(x_t | x_{t-1})$, and a learnable inverse process, denoted as $p_{\theta}(x_{t-1} | x_t)$. The forward process gradually transforms data from the target distribution $p(x_0)$ into a Gaussian distribution by adding noise, while the inverse process generates samples by transforming noise into $q(x_0)$, as shown in Fig. \ref{Fig.4}.

\begin{figure}
	\centering
	\includegraphics[width=0.5\linewidth]{"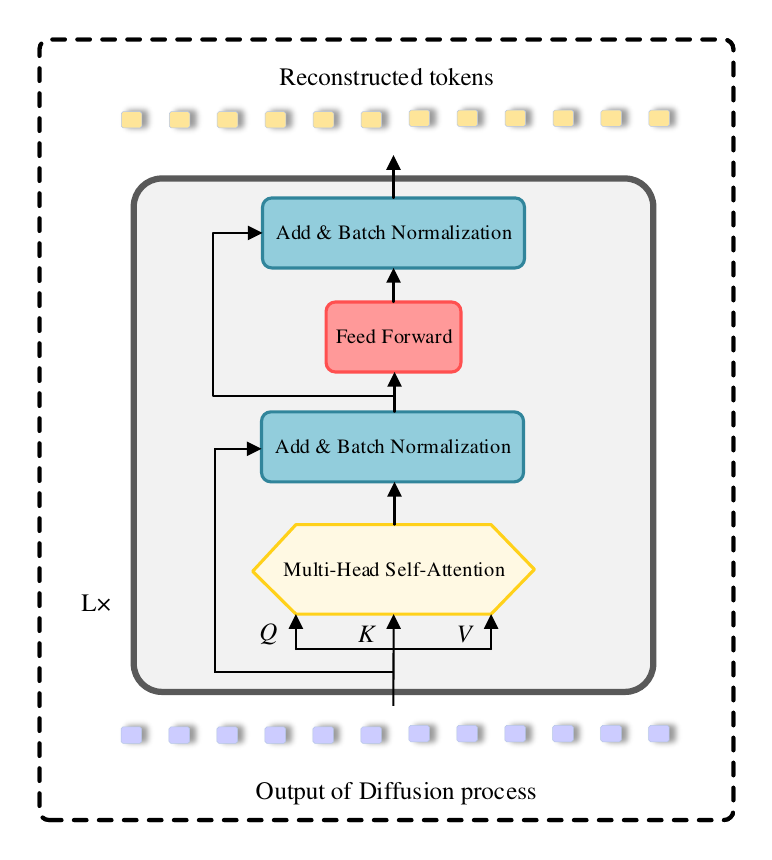"}
	\caption{In the DDT, we utilize a standard Transformer encoder block as the internal neural network to learn the reverse process. For simplicity, we denote the query, key, and value of interest as Q, K, and V.}
	\label{Fig.5}
\end{figure}

The forward process of the Denoising Diffusion Model is a non-homogeneous Markov noise process. In this process, at each time step $t$, Gaussian noise with specific mean and variance is added to the data. This dynamic process can be described as follows:

\begin{equation}
	q(\left.x_1,\ldots,x_T\mid x_0\right)=\prod\limits_{t=1}^T q(\left.x_t\mid x_{t-1}\right)
	\label{(1)}
\end{equation}
where $q(x_t\mid x_{t-1})=\mathcal{N}\big(x_t\mid x_{t-1}\sqrt{1-\beta_t},\beta_t\mathbf{I}\big)$. The noise added at each step is determined by a variance table $\beta_t\in(0,1), t=1,...,T$. It can be defined as a small linear sequence. Based on the updates of $\beta$ as described in Ho et al. \cite{ho2020denoising}, we set $\beta_1=10^{-4}$ and linearly increase $\beta_t$ up to 0.02. By computing available $x_t=\sqrt{1-\beta_t}x_{t-1}+\sqrt{\beta_t}\epsilon_{t-1}$, for $\epsilon_{t-1}{}{\sim}N(0,\mathbf{I})$. The denoising process $p_\theta$ is learned by optimizing the model parameters $\theta$ and is defined as follows:

\begin{equation}
	p_\theta(\begin{matrix}x_0,\dots,x_{t-1}\mid x_T\end{matrix})=p(x_T)\prod\limits_{t=1}^Tp_\theta(\begin{matrix}x_{t-1}\mid x_t\end{matrix})
	\label{(2)}
\end{equation}
where $p_\theta(x_{t-1}\mid x_t)=\mathcal{N}\big(x_{t-1}\mid\mu_\theta(x_t,t),\tilde{\beta}_t\text{I}\big)$. As demonstrated by Ho et al. \cite{ho2020denoising}, it is possible to train the inverse process using the following objective function:

\begin{equation}
	L=min_{\theta}\mathbb{E}_{x_{0}\sim\mathcal{D},\epsilon\sim\mathcal{N}(0,I),t\sim u(1,T)}||\epsilon-\epsilon_{\theta}\Big(\sqrt{\alpha_{t}}x_{0}+(1-\alpha_{t})\epsilon,t\Big)||^{2}
	\label{(9)}
\end{equation}
where $\mathcal{D}$ refers to the data distribution, and $\epsilon_\theta(x_t,t)$ is parameterized using a neural network. For brevity, $L$ refers to the $\mathcal{L}_\mathrm{simple}$ in Ho et al. \cite{ho2020denoising}. We have replaced the U-Net with Transformer as the neural network architecture inside the Denoising Diffusion Model to better handle deep local features and long-term dependencies within time series data.
replacing the U-Net
The original Denoising Diffusion Model used a U-Net as the internal neural network to learn the reverse process. In this paper, the convolutional and pooling layers of the U-Net in the Denoising Diffusion Model have been replaced with the encoder portion of the Transformer. This modification helps us capture both local and global relationships within the time series data. The core feature of the Transformer model is its self-attention mechanism, which allows the model to assign different weights to each element in the input sequence, taking into account information from other elements simultaneously. This design enables the model to handle long-range dependencies more effectively and benefits from the parallel computation.
Furthermore, the multi-head attention mechanism allows the model to focus on different subspaces, enhancing its representational capacity. Feed-forward neural networks are used to process further intermediate representations, and residual connections and layer normalization techniques help mitigate the problem of gradient vanishing, thereby speeding up the training process. Fig. \ref{Fig.5} illustrates the architecture of the Transformer encoder portion.

The implementation process of the DDT is illustrated in Algorithm \ref{alg:denoising} as follows:
	
\begin{algorithm}
	\caption{Denoising Diffusion Transformer}
	\label{alg:denoising}
	\textbf{Initialization:}
	\begin{algorithmic}
		\State $x_0 \sim q(x_0), \epsilon \sim N(0,I), \beta_t, \alpha_t = 1 - \beta_t, \bar{\alpha}_t = \prod_{i=0}^T \alpha_i$
		\State $t \sim \text{Uniform}(\{1, \ldots, T\})$
	\end{algorithmic}
	\textbf{Repeat}
	\begin{algorithmic}
		\State Take gradient descent step on $\nabla_\theta\parallel\boldsymbol{\epsilon}-\boldsymbol{\epsilon}_{\theta}(\sqrt{\bar{\alpha}_{t}}x_{0}+\sqrt{1-\bar{\alpha}_{t}}\boldsymbol{\epsilon},t)\parallel^{2}$
	\end{algorithmic}
	$\mathbf{until}$ converged\\
	\textbf{Noise: }$x_T \sim \mathcal{N}(0,I)$
	\begin{algorithmic}
		\For{$t = T,\ldots,1$}
		\If{$t > 1$}
		\State $z \sim \mathcal{N}(0,I)$
		\Else
		\State $z = 0$
		\EndIf
		\State $x_{t-1} = \frac{1}{\sqrt{\alpha_t}}\left(x_t - \frac{\beta_t}{\sqrt{1-\bar{\alpha}_t}}\epsilon_{\theta}(x_t,t)\right) + \sigma_t z$
		\EndFor
		\State \textbf{Return} $x_0$
	\end{algorithmic}
\end{algorithm}

\section{Experiment}

In this section, we first describe the configurations and properties related to the experiments. Subsequently, we perform benchmark tests on the DDMT using five real-world datasets, and evaluate it against several competitive baselines. The source code of the model can be obtained from \href{https://github.com/yangchaocheng/DDMT.git}{https://github.com/yangchaocheng/DDMT.git}.

\subsection{Experiment Settings}
\subsubsection{Datasets}
We collected five real-world datasets to test our method. Here is a description of the five experimental datasets:

(1)\textbf{Server Machine Dataset(SMD)}\cite{su2019robust}: The SMD dataset was collected from a large internet company and spans five weeks. It consists of multivariate time series data with 38 dimensions.

(2)\textbf{Pooled Server Metrics(PSM)}\cite{abdulaal2021practical}:  The PSM was collected internally from multiple application server nodes at eBay. It contains data from 26 dimensions.

(3)\textbf{Mars Science Laboratory rover(MSL)} and \textbf{Soil Moisture Active Passive satellite(SMAP)}\cite{hundman2018detecting}: The MSL and SMAP are public datasets provided by NASA. The MSL dataset consists of 55 dimensions, while the SMAP dataset has 25 dimensions. These datasets include telemetry anomaly data reported by the spacecraft detection system called Incident Surprise Anomaly (ISA).

(4)\textbf{Secure Water Treatment(SWaT)}\cite{mathur2016swat}: The SWaT dataset comprises data from 51 sensors and originates from a continuously operating critical infrastructure system, specifically a water treatment facility. The information on all datasets is shown in Table \ref{tab:dataset}.

\begin{table}
	\centering
	\caption{Dataset Information. Anomalies represents the proportion of abnormal in the entire dataset.}
	\setlength{\tabcolsep}{13pt}
	\label{tab:dataset}
	\begin{tabularx}{\textwidth}{cccccc}
		\toprule
		\textbf{Name} & \textbf{Train} & \textbf{Test} & \textbf{Dimension}  & \textbf{Anomalies} & \textbf{Applications} \\
		\midrule
		SMD & 708,405 & 708,420 & 38 & 0.042 & Server \\
		PSM & 132,481 & 87,841 & 25 & 0.278 & Server \\
		MSL & 58,317 & 73,729 & 55 & 0.105 & Space \\
		SMAP & 135,183 & 427,617 & 25 & 0.128 & Space \\
		SWaT & 495,000 & 449,919 & 51 & 0.121 & Water \\
		\bottomrule
	\end{tabularx}
\end{table}

\subsubsection{Implementation Details}

Based on the improved approach by Shen et al. \cite{shen2020timeseries}, we adopted non-overlapping sliding windows to obtain a set of subsequences. For all the datasets, we set the sliding window size to 100. If the anomaly score exceeds a certain threshold, the corresponding time point is labeled as an anomaly. We determined the threshold based on the proportion of anomalies in the validation dataset, as shown in Table \ref{tab:dataset}. We employed a widely used adjustment strategy \cite{xu2018unsupervised,su2019robust,shen2020timeseries}, where if a time point within a consecutive period of anomalies is detected, all anomalies within that period are considered correctly detected. This strategy is reasonable because the observed time points of anomalies trigger alerts and further draw attention to the entire time period in practical applications.

We utilized the SGD optimizer with an initial learning rate of 1e-4, a mask scale of 5, and 500 diffusion steps during the denoising process. The training batch size was 256, and the training process was stopped early within 10 epochs. All experiments were conducted using a single NVIDIA RTX 3090 24 GB GPU with implementation in PyTorch 1.13.

\subsubsection{Evaluation Metrics}

In this study, we employ the following formulas to calculate the precision, recall, and F1-score for evaluating the performance of our anomaly detection method. Precision is defined as:

\begin{equation}
	Precision=\frac{TP}{TP+FP}
	\label{(10)}
\end{equation}
where $TP$ represents true positives (the number of samples correctly predicted as anomalies) and $FP$ represents false positives (the number of samples incorrectly predicted as anomalies). Recall is defined as:

\begin{equation}
	Recall=\frac{TP}{TP+FN}
	\label{(11)}
\end{equation}
where $FN$ represents false negatives (the number of samples incorrectly predicted as normal). $F_1$ is defined as:

\begin{equation}
	F_1=\frac{2*Precision*Recall}{Precision+Recall}
	\label{(12)}
\end{equation}

Precision measures the model's accuracy in predicting samples as anomalies, while recall assesses the model's ability to cover actual anomalies. The F1-score balances precision and recall. By computing precision, recall, and F1-score, we can comprehensively evaluate the performance of anomaly detection methods, quantifying the model's accuracy, coverage, and overall performance in anomaly detection tasks. Finally, we employed statistical methods to assess performance differences among various models. We used the Independent Samples t-test to calculate the F1-score differences across multiple datasets for different models and obtained corresponding p values.

\subsection{Performance}

To evaluate the performance of DDMT, we conducted extensive comparisons with 15 well-performing baseline models, including machine learning-based TSAD models: Isolation Forest \cite{liu2008isolation}, LOF \cite{breunig2000lof}, One-class Support Vector Machines (OC-SVM) \cite{ma2003time}, Deep Autoencoding Gaussian Mixture Model (DAGMM) \cite{zong2018deep}, and  Temporal Hierarchical One-Class (THOC) \cite{shen2020timeseries}. Deep learning-based anomaly detection models: Long Short-Term Memory-based Variational AutoEncoder (LSTM-VAE) \cite{park2018multimodal}, OmniAnomaly \cite{su2019robust}, UnSupervised Anomaly Detection (USAD) \cite{audibert2020usad}, InterFusion \cite{li2021multivariate}, Anomaly Transformer(AT) \cite{xu2021anomaly}, Transformer-based Anomaly Detection model (TranAD) \cite{tuli2022tranad}, TimesNet \cite{wu2022timesnet}, Dual-Channel Feature Fusion (DCFF-MTAD) \cite{xu2023dcff}, Dilated Convolutional Transformer-based GAN (DCT-GAN) \cite{li2021dct}, and Multiscale wavElet Graph AE (MEGA) \cite{wang2022multiscale}. Table \ref{tab:performance} presents a comprehensive performance comparison between our model and other baseline models. The best results are indicated in bold black, while the second-best results are represented with underlines.

\begin{table}
	\centering
	\caption{Performance comparison of different models. The \textbf{P}, \textbf{R}, and \textbf{F1} represent  the precision, recall, and F1-score (as \%), respectively. The F1-score is the harmonic mean of precision and recall. Higher values for these three metrics indicate better performance.}
	\label{tab:performance}
	\setlength{\tabcolsep}{3pt}
	\small
	\begin{tabularx}{1\textwidth}{@{} c| *{3}{c} | *{3}{c} | *{3}{c} | *{3}{c} | *{3}{c} @{}}
		\toprule
		\multirow{3}{*}{\textbf{Model}} & \multicolumn{3}{c}{\textbf{SMD}} & \multicolumn{3}{c}{\textbf{MSL}} & \multicolumn{3}{c}{\textbf{SMAP}} & \multicolumn{3}{c}{\textbf{SWaT}} & \multicolumn{3}{c}{\textbf{PSM}} \\
		\cmidrule(lr){2-4} \cmidrule(lr){5-7} \cmidrule(lr){8-10} \cmidrule(lr){11-13} \cmidrule(lr){14-16}
		& \textbf{P} & \textbf{R} & \textbf{F1} & \textbf{P} & \textbf{R} & \textbf{F1} & \textbf{P} & \textbf{R} & \textbf{F1} & \textbf{P} & \textbf{R} & \textbf{F1} & \textbf{P} & \textbf{R} & \textbf{F1} \\
		\midrule
		IsolationForest & 40.58 & 78.92 & 53.60 & 58.18 & 82.15 & 68.12 & 58.12 & 62.36 & 60.17 & 48.69 & 42.98 & 45.66 & 78.26 & 89.35 & 83.44 \\
		LOF & 59.45 & 36.62 & 45.32 & 53.25 & 82.38 & 64.69 & 58.62 & 59.89 & 59.25 & 76.67 & 70.15 & 73.27 & 60.68 & 86.37 & 71.28 \\
		OC-SVM & 48.12 & 78.15 & 59.56 & 56.37 & 89.12 & 69.06 & 56.34 & 62.36 & 59.20 & 48.65 & 46.13 & 47.36 & 68.35 & 78.26 & 72.97 \\
		DAGMM & 65.15 & 52.64 & 58.23 & 83.78 & 66.46 & 74.12 & 89.35 & 53.26 & 66.74 & 86.45 & 60.34 & 71.07 & 89.15 & 68.12 & 77.23 \\
		THOC & 81.45 & 88.26 & 84.72 & 85.12 & 86.69 & 85.90 & 90.18 & 88.65 & 89.41 & 86.87 & 83.64 & 85.22 & 86.34 & 85.61 & 85.97 \\
		LSTM-VAE & 78.12 & 85.62 & 81.70 & 86.23 & 78.61 & 82.24 & 90.12 & 70.25 & 78.95 & 79.23 & 86.25 & 82.59 & 76.15 & 86.25 & 80.89 \\
		OmniAnomaly & 80.12 & 89.36 & 84.49 & 88.56 & 83.12 & 85.75 & 93.58 & 80.12 & 86.33 & 84.36 & 82.39 & 83.36 & 85.69 & 76.39 & 80.77 \\
		USAD & 90.35 & 96.17 & 93.17 & 83.26 & 93.28 & 87.99 & 75.68 & 96.25 & 84.73 & 95.32 & 76.39 & 84.81 & 76.25 & 97.36 & 85.52 \\
		InterFusion & 85.18 & 80.45 & 82.75 & 85.38 & 90.78 & 88.00 & 90.89 & 86.28 & 88.53 & 82.39 & 86.34 & 84.32 & 80.68 & 86.34 & 83.41 \\
		AT & 89.92 & 94.67 & 92.23 & 93.76 & 95.08 & 94.41 & 92.45 & 98.64 & \uline{95.44} & 92.38 & 96.40 & \uline{94.34} & 97.76 & 98.28 & \uline{98.01} \\
		TranAD & 90.36 & 97.85 & \uline{93.96} & 90.38 & 99.99 & \uline{94.94} & 85.12 & 95.63 & 90.07 & 96.39 & 73.68 & 83.52 & 89.90 & 96.38 & 93.03 \\
		TimesNet & 87.76 & 82.63 & 85.12 & 82.97 & 85.42 & 84.18 & 91.50 & 57.80 & 70.85 & 88.31 & 96.24 & 92.10 & 98.22 & 92.21 & 95.21 \\
		DCFF-MTAD & 86.28 & 95.10 & 90.48 & 90.16 & 92.64 & 91.38 & 98.36 & 83.96 & 90.59 & 95.68 & 89.12 & 92.28 & 94.39 & 92.98 & 93.68 \\
		DCT-GAN & 73.53 & 81.25 & 77.20 & 80.36 & 70.23 & 74.95 & 89.28 & 76.67 & 82.50 & 80.68 & 81.62 & 81.15 & 81.85 & 76.98 & 79.34 \\
		MEGA & 96.58 & 96.28 & \textbf{96.43} & 93.65 & 91.64 & 92.63 & 88.36 & 83.64 & 85.94 & 90.16 & 94.31 & 92.19 & 94.25 & 95.01 & 94.63 \\
		\midrule
		OURS & 88.07 & 81.50 & 84.66 & 92.67 & 98.07 & \textbf{95.30} & 94.08 & 99.08 & \textbf{96.51} & 96.56 & 99.78 & \textbf{97.24} & 97.92 & 98.40 & \textbf{98.16} \\
		\bottomrule
	\end{tabularx}
\end{table}

Table \ref{tab:performance} presents the accuracy, recall, and F1-score of all datasets on the DDMT and baseline models. From the table, it can be observed that in the machine learning-based anomaly detection models, THOC achieved superior results, attributed to its use of dilated recursive neural networks with skip connections to capture temporal dynamics at multiple scales and the incorporation of a self-supervised task in the time domain \cite{shen2020timeseries}.On the other hand, the performance of DCT-GAN in the deep learning-based anomaly detection models is relatively poor, which may be primarily due to the utilization of PCA to transform multidimensional data into one-dimensional data for anomaly detection, which hinders the exploration of deep features and correlations among different dimensions in multidimensional data \cite{li2021dct}. In addition to the DCT-GAN model, deep learning-based models generally outperform machine learning-based models in F1-score. Anomaly Transformer and TranAD achieved second-best results in five datasets twice and three times, respectively, while the MEGA model achieved the best result once.

Furthermore, we observed that DDMT demonstrated slightly better performance in F1-score compared to other baseline models, surpassing the best baseline results by 0.36\%, 1.07\%, 2.90\%, and 0.15\% in the MSL, SMAP, SWaT, and PSM datasets, respectively. For the SMD dataset, the MEGA model obtained the highest F1-score of 96.43.

\begin{figure}
	\centering
	\includegraphics[width=0.8\linewidth]{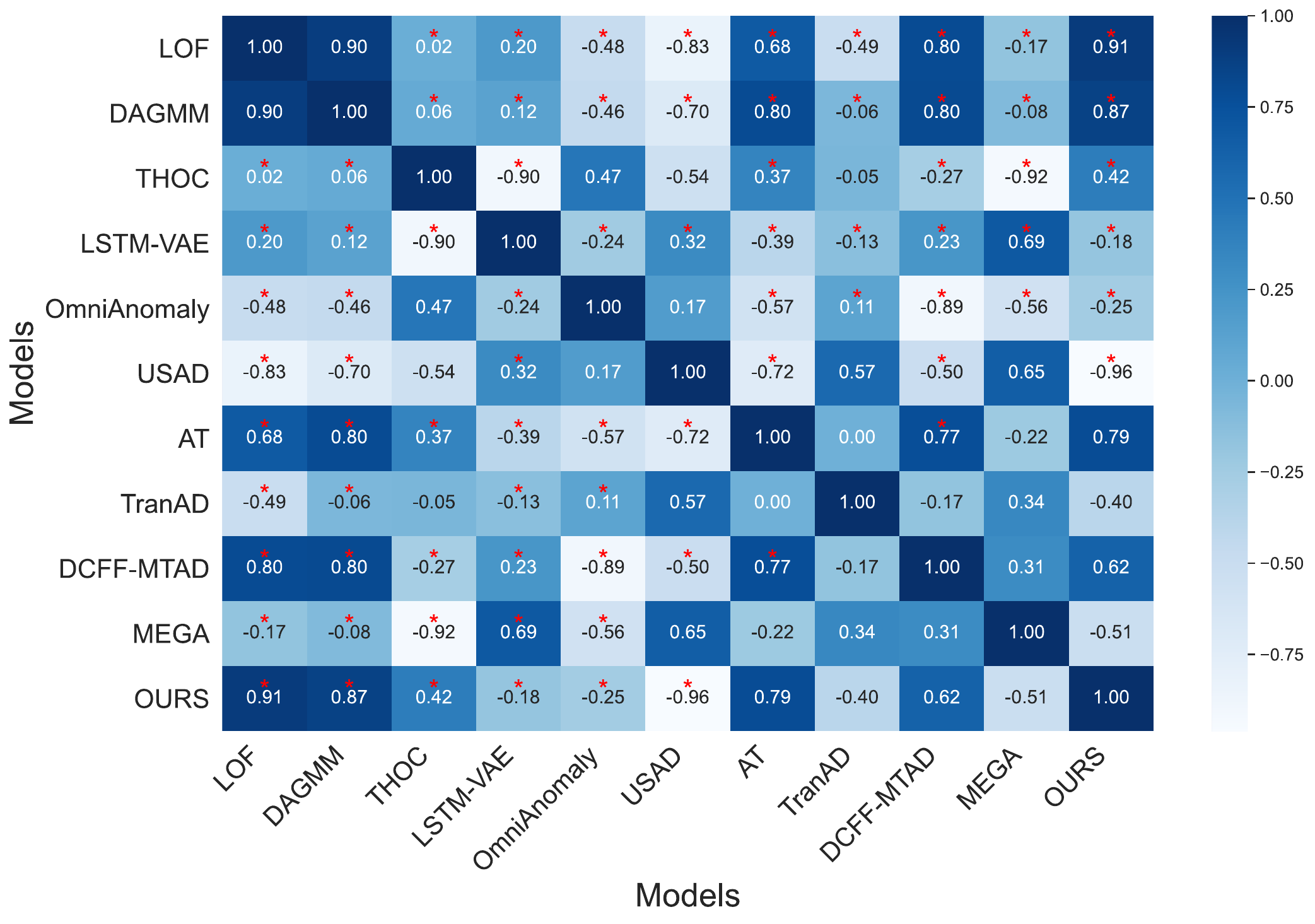}
	\caption{F1-Score Correlation Heatmap. The depth of the square color represents the level of correlation, and the red asterisk represents a p-value less than 0.05, indicating a significant difference.}
	\label{Fig.6}
\end{figure}

To assess the differences between DDMT and other representative baseline models in both machine learning and deep learning-based anomaly detection, we selected 10 models for comparison. We used a t-test as the statistical method to calculate the p values for the differences in F1 scores between each pair of models. Following the convention, we set the significance level at 0.05. When the p value was less than 0.05, we consider the difference to be significant, indicating a notable distinction in F1 scores between different models. The results were then visualized in a heatmap (Fig. \ref{Fig.6}) to provide a more intuitive representation of the variations among the models.

In Fig. \ref{Fig.6}, the color intensity represents the correlation of the F1 scores, with lighter colors indicating a higher correlation. Additionally, we added corresponding p value markers to the heatmap to highlight the significance of the differences. Specifically, when the p value is less than 0.05, we used red asterisks to denote significant differences. From the heatmap, it can be observed that DDMT exhibits significant differences and holds statistical significance when compared to the LOF, DAGMM, THOC, LSTM-VAE, OmniAnomaly, and USAD models. These results demonstrate the distinctiveness of DDMT concerning F1 scores compared to the selected baseline models.

\subsection{Ablation Experiments}

The experimental results above indicate that DDMT is feasible for anomaly detection in multivariate time series data. To investigate the effectiveness of each component of our method, we gradually removed the constituents of the model while keeping the settings unchanged. We observed the performance changes of the model on each dataset accordingly. Table \ref{tab:comparison} presents the experimental results comparison after eliminating various modules of the model.

\begin{table}
	\centering
	\caption{Ablation study of ADNM and DDT on the different datasets.}
	\label{tab:comparison}
	\begin{tabular}{lcccc}
		\toprule
		Dataset & Transformer & W/O ADNM & W/O DDT & Our \\
		\midrule
		SMD     & 79.56       & 83.16       & 76.73       & 84.66 \\
		MSL     & 78.68       & 91.26       & 87.26       & 95.30 \\
		SMAP    & 69.70       & 93.02       & 90.86       & 96.51 \\
		SWaT    & 80.37       & 90.12       & 92.16       & 97.24 \\
		PSM     & 76.07       & 95.36       & 90.93       & 98.16 \\
		\midrule
		Average & 76.88       & 90.58       & 87.59       & 94.37 \\
		\bottomrule
	\end{tabular}
\end{table}

\textbf{Our Without DDT:} In this study, we investigated the impact of the DDT module on the results by removing both the forward and backward processes of the Denoising Diffusion Model. From the table, it can be observed that after removing the DDT module, the model's performance decreased. The DDT module contributed to an average F1-score improvement rate of 7.18\% (87.19\%→94.37\%). The results demonstrate that the DDT module is highly effective and essential for the model's overall performance.

\textbf{Our Without ADNM:} We further explored the impact of the ADNM module. In this study, we replaced the ADNM with the original self-attention module to investigate whether ADNM could enhance the anomaly detection performance of the model. From the table, it is evident that the ADNM module significantly improves the anomaly detection effectiveness. The proposed ADNM with the adaptive dynamic neighbor mask mechanism outperforms the original self-attention model with an average F1-score improvement of 3.99\% (90.38\%→94.37\%). The results indicate that avoiding information leakage of each token and its neighbors during sequence reconstruction is crucial, and the masking mechanism in ADNM helps mitigate the identical shortcut problem during sequence reconstruction.

\textbf{Baseline:} To understand the effectiveness of the  proposed method, we also compared it with a Transformer model as the baseline. In the Transformer model, the traditional self-attention mechanism was used to capture deep-level features of the time series by computing the attention map between all tokens for sequence reconstruction. The experimental results show that the average F1-score of the DDMT model was improved by 16.95\% (77.42\%→94.37\%) compared to the Transformer model, demonstrating a significant superiority of the DDMT model over the Transformer model. In other words, the ADNM and DDT components in the DDMT model substantially enhance the anomaly detection performance.

\subsection{Analysis DDMT}

In this section, we investigated the impact of various hyperparameters on the performance of the DDMT model. We primarily analyzed the effects of the window size on the results, as well as the influence of the mask scale in the ADNM module and the diffusion steps $\mathcal{S}$ in the DDT denoising process on the experimental outcomes.

\subsubsection{Analysis of Window Size}

\begin{figure}
	\centering
	\includegraphics[width=0.8\linewidth]{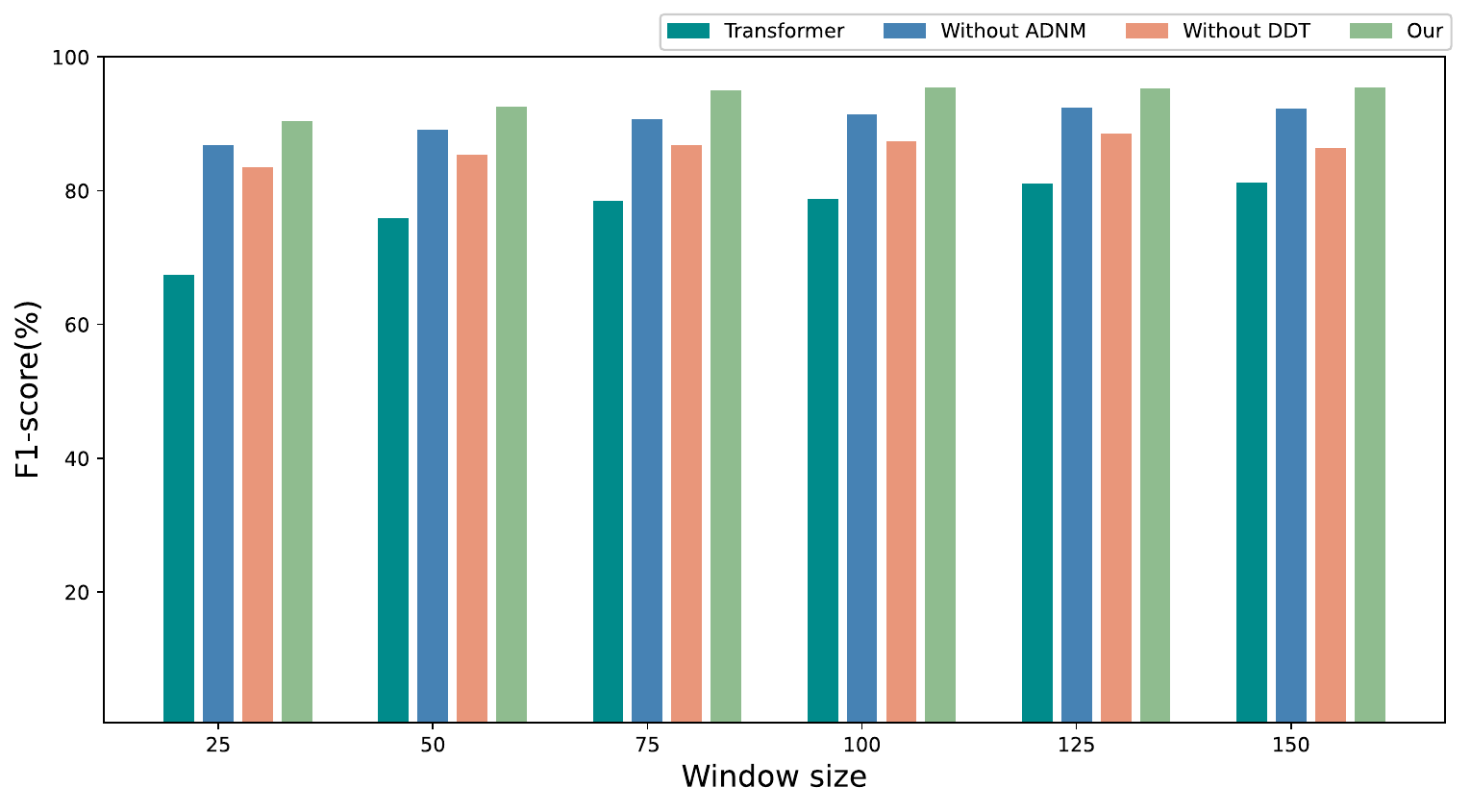}
	\caption{The F1 comparison of different window size.}
	\label{Fig.7}
\end{figure}

The sliding window size is an important parameter influencing the model's performance. For the given MSL dataset, we conducted experiments with four different models using various window sizes. The graph illustrates the impact of window size on the F1-scores of the different models, where each colored bar represents a specific model in the ablation experiments. As shown in Fig. \ref{Fig.7}, it can be observed that despite the differences in model architectures, the anomaly detection performance of all models gradually improves as the window size increases. The experiments demonstrate that the window size significantly affects the final results and should not be overlooked.

\subsubsection{Sensitivity to Mask Scale}

The mask scale primarily reflects the level of attention the model pays to each token and its neighboring information during time series reconstruction. During sequence reconstruction, the model should not overly rely on the information of each token and its neighbors, as subtle anomalies might be directly mapped onto the reconstructed time series, making it challenging to distinguish whether they are anomalies based on the differences between the original and reconstructed time series. However, if the sequence information is excessively masked, it can hinder the model's ability to capture deep-level features of the time series, which is essential for effective reconstruction. Therefore, selecting an appropriate mask scale is crucial for the time series reconstruction process.

\begin{figure}
	\centering
	\includegraphics[width=0.7\linewidth]{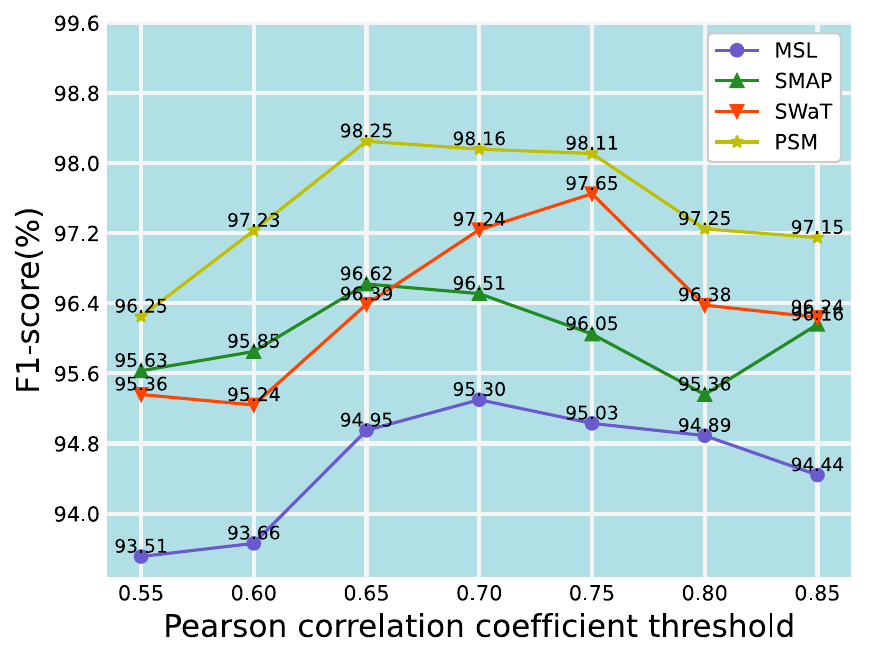}
	\caption{The performance of the model based on varying mask scale. The Pearson correlation coefficient threshold is used to determine the number of neighboring nodes that should be masked. A higher correlation coefficient results in a larger mask scale, indicating a greater number of neighboring nodes that need to be masked.}
	\label{Fig.8}
\end{figure}

To clearly observe the sensitivity of F1-score to the mask scale, we conducted experiments on four datasets: MSL, SMAP, SWaT, and PSM. We controlled the size of the mask scale using Pearson correlation coefficient thresholds, where a higher correlation coefficient threshold results in a larger mask scale. The graph shows the F1-score performance for each dataset at different Pearson correlation coefficient thresholds. As shown in Fig. \ref{Fig.8}, it is evident that as the Pearson correlation coefficient threshold increases, the model's anomaly detection performance gradually improves. However, when the Pearson correlation coefficient threshold reaches 0.7, the model's performance starts to decline. The experiments demonstrate that masking some degree of self and neighboring information can enhance the model's anomaly detection performance. However, if the mask scale becomes too large, it may overlook some deep-seated features within the time series, which is detrimental to time series reconstruction and, consequently, affects anomaly detection performance.

\subsubsection{Effect of Diffusion Steps $\mathcal{S}$}

In this section, we considered the influence of the diffusion steps $\mathcal{S}$ in the iterative denoising process of the Denoising Diffusion Model on the P, R, F1-score, and the average time needed for the anomaly detection process. The diffusion steps $\mathcal{S}$ in the denoising process are crucial hyperparameters. If the value of $\mathcal{S}$ is too small, the model may not fully reduce noise and capture fine-grained information in the time series. On the other hand, if the value of $\mathcal{S}$ is too large, the model may overly denoise the data, reducing the model's robustness and requiring excessive computational resources and time to complete the iteration process of the model. Therefore, selecting an appropriate value of $\mathcal{S}$ is essential to achieve effective denoising and enhance the anomaly detection performance of the model.

\begin{figure}
	\centering
	\includegraphics[width=0.7\linewidth]{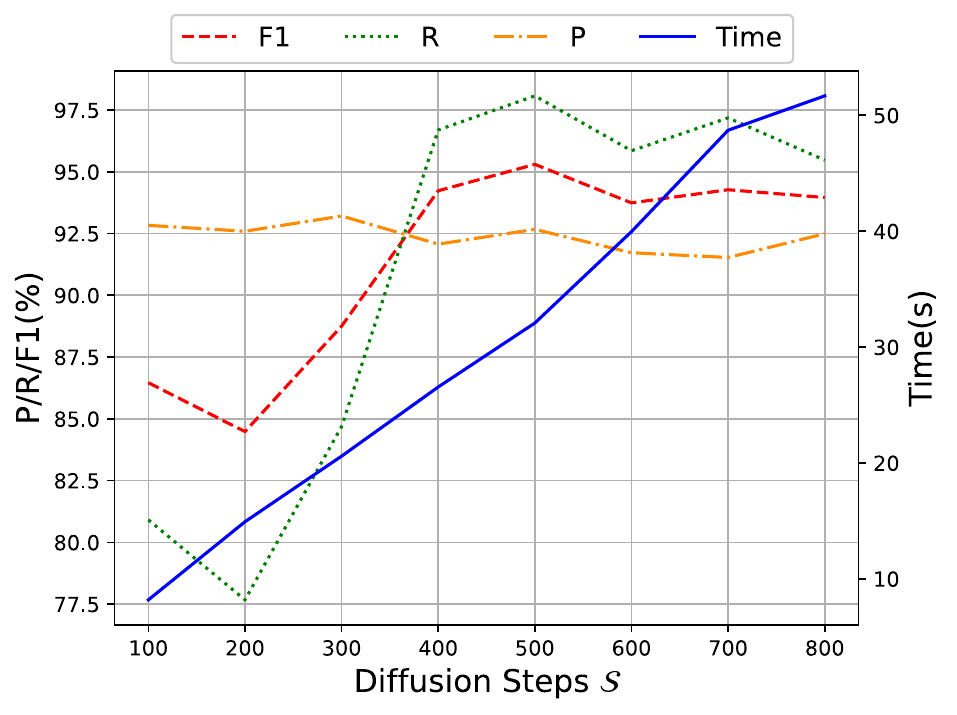}
	\caption{The \textbf{P}, \textbf{R}, \textbf{F1} and anomaly detection time with diffusion steps $\mathcal{S}$. The \textbf{P}, \textbf{R}, and \textbf{F1} represent  the precision, recall, and F1-score (as \%), respectively.}
	\label{Fig.9}
\end{figure}

As shown in Fig. \ref{Fig.9}, we can observe that as the value of $\mathcal{S}$ increases, the anomaly detection performance of the model gradually improves. However, when $\mathcal{S}$ reaches 500, the model’s performance shows little further improvement while the training time increases rapidly. The experiments demonstrate that although increasing the diffusion steps $\mathcal{S}$ in the denoising process enhances the anomaly detection effectiveness, it also requires more computational resources. Therefore, selecting an appropriate value of $\mathcal{S}$ is crucial to finding a balance between model performance and computational resources for effective anomaly detection.

\section{Conclusion}

This paper investigates anomaly detection in multidimensional time series data. In contrast to previous approaches, we combine the Denoising Diffusion Model with the Transformer model, introducing a novel TSAD model called DDT and designing a novel mask mechanism ADNM. DDT combines the Denoising Diffusion Model and Transformer, utilizing the Transformer as the internal neural network structure to effectively capture both global and local dependencies in time series data. As a state-of-the-art generative model, the Denoising Diffusion Model is employed for time series generation through a forward diffusion process and a learnable reverse process. ADNM dynamically adjusts masks, masking information from anomalous nodes and their neighboring nodes in the time series, alleviating the issue of WIM during sequence reconstruction and enhancing the model's robustness. We conducted extensive experiments on five publicly available datasets, comparing our proposed model with existing anomaly detection models, and demonstrated the effectiveness of our approach. Ablation studies confirmed the effectiveness of each component within DDMT and their ability to handle noise and capture subtle anomalies.

Similarly, DDMT also exhibits certain limitations. Our future work will address these shortcomings from two perspectives:

1. \textbf{Incorporating Prior Knowledge:} DDMT currently performs end-to-end training on datasets without incorporating prior knowledge, such as specific holidays or significant events. This limitation may result in false positives or false negatives in certain situations. Introducing domain-specific expert knowledge during the training process often benefits the anomaly detection performance of the model.

2. \textbf{Reducing Computational Resource Requirements:} The iterative process of the Denoising Diffusion Model consumes substantial computational resources. In resource-constrained environments, it might be challenging to detect anomalies within a short timeframe. Therefore, we aim to improve the model's architecture to reduce the running time and enhance the utilization of computational resources.

\bibliographystyle{unsrt}
\bibliography{MyCollection} 

\end{document}